\documentclass[a4paper]{article}
\usepackage{INTERSPEECH2019,amsmath,graphicx,hyperref,tikz}
\usetikzlibrary{decorations.pathreplacing, positioning, arrows.meta}

\usepackage[T1]{fontenc}
\usepackage[english]{babel}
\usepackage[utf8]{inputenc}

\usepackage{multirow}
\usepackage{microtype}
\usepackage{booktabs}
\usepackage{todonotes}
\usepackage{mathtools}
\usepackage{hyperref}

\usepackage{tikz}

\newcommand{\blank}{\emptyset}
\newcommand{\spc}{\:\:}

\renewcommand{\L}{\mathcal{L}}

\title{Towards using context-dependent symbols in CTC\\without state-tying decision trees}

\name{Jan Chorowski, Adrian Lancucki, Bartosz Kostka, Michal Zapotoczny\thanks{The authors thank Polish National Science Center for funding under the Sonata 2014/15/D/ST6/04402 grant and to the PLGrid project for computational resources on the Prometheus cluster.}}
\address{University of Wrocław, Poland}
\email{jch@cs.uni.wroc.pl}

\begin{document}
\maketitle
\begin{abstract}
Deep neural acoustic models benefit from context-dependent (CD) modeling of output symbols.
We consider direct training of CTC networks with CD outputs,
and identify two issues. The first one is frame-level normalization of probabilities in CTC,
which induces strong language modeling behavior that leads to overfitting and interference with external language models.
The second one is poor generalization in the presence of numerous lexical units like triphones or tri-chars.
We mitigate the former with utterance-level normalization of probabilities.
The latter typically requires reducing the CD symbol inventory with
state-tying decision trees,
which have to be transferred from classical GMM-HMM systems.
We replace the trees with a CD symbol embedding network, which saves parameters and ensures
generalization to unseen and undersampled CD symbols.
The embedding network is trained together with the rest of the
acoustic model and removes
one of the last cases in which neural systems have to be bootstrapped from GMM-HMM ones.
\end{abstract}

\section{Introduction}
\label{sec:intro}
Acoustic models built with end-to-end trained deep neural networks are general enough to read raw waveforms \cite{hoshen2015speech} and output characters \cite{amodei2016deep}, bypassing steps of feature extraction and preparation of pronunciation lexicons. However, even with abundant training data, the performance of end-to-end networks can be improved by introducing elements from classical GMM-HMM  models, most notably the concept of context-dependent (CD) output symbols. To limit the number of considered states, these are clustered using state-tying decision trees \cite{senior2015context,hadian18end-to-end}, often taken from GMM-HMM models, 
making for better representations of CD symbols which are rare or absent in the training data.

In this work we consider Connectionist Temporal Classification (CTC) \cite{graves2006connectionist} networks with context dependency: for each frame of acoustic features, the network predicts CD targets (phonemes or characters in context).
To the advantage of CTC networks, the training procedure can be started
on unaligned transcripts and does not require bootstrapping from a %
GMM-HMM system.
To preserve these properties, we propose to compute the representations of CD output symbols 
with %
an embedding network, which is a neural analogue
to a state-tying decision tree.
The embedding network naturally exploits similarities between contexts,
and generalizes to previously unseen ones.

We address the challenges posed by using CD symbols with CTC.
We show that using CD symbols improves over a vanilla CTC solution,
and that the best results are obtained with the formulation of CTC loss
more similar to %
the Lattice-Free MMI methods.
During training, it normalizes probabilities on the transcript level
rather than on the frame level.

\section{Background}
\subsection{GMM-HMMs with Decision Trees}
The classical GMM-HMM acoustic model assumes that each speech segment (typically a 25ms long acoustic frame) is emitted from a single Gaussian mixture specified by the hidden state of the HMM \cite{rabiner1989tutorial,gales2007application}. This simple emission model assumes similarity of frames emitted from the same state, and dissimilarity to the remaining frames.
It is achieved by careful modeling of acoustic phenomena.
Each phoneme is partitioned into three sub-phonemic states, and
context-dependent changes in pronunciation (e.g., voicing)
are handled by differentiating between the emissions of the same phoneme from different contexts. 

Particularly popular are triphones, i.e., phonemes considered in their left and right contexts.
Larger units like quinphones are possible as well.
To cope with the large amount of possible CD symbols the emission GMMs are shared between contexts, with the mapping performed by state-tying decision trees \cite{young1994tree}. Typically, such trees map each CD symbol to one of a few thousand GMMs (also called tied states).

In a hybrid DNN-HMM model the Gaussian mixture representation of speech frames is replaced with a deep neural network that is tasked with predicting the HMM state aligned with each frame. In principle, the network can become invariant to many acoustic phenomena and it is no longer necessary to model them explicitly. Indeed, little to no gains are reported for dividing phonemes into subphonemic states \cite{senior2015context}. However, explicit modeling of the context still improves recognition accuracy \cite{senior2015context,hadian18end-to-end}. For this reason, a typical DNN-HMM system employs neural networks that predict the tied states of a previously trained GMM-HMM model. Consequently, the application of neural networks still depends on decision trees that map CD symbols to individual network outputs.

\subsection{End-to-end approaches to speech recognition}
End-to-end systems can be trained from scratch without a dependency on a previously built GMM-HMM system.
They employ cost functions able to establish
an alignment between the sequence of acoustic frame features and the elements of the target transcript.
This can be accomplished in two ways. First, a sequence-level cost function can be applied to an HMM topology.
Examples of such models are CTC \cite{graves2006connectionist,miao2015eesen}, Lattice-Free Maximal Mutual Information (LF-MMI) \cite{povey_purely_2016,hadian18end-to-end} and Graph Transformer Networks (GTN) \cite{lecun_gradientbased_1998,collobert_wav2letter_2016}. 
A second family of models
replaces HMMs with a neural attention mechanism \cite{chorowski_attentionbased_2015a, chan_listen_2015}.
In this contribution we focus on CTC due to its popularity and wide adoption in recent ASR projects
such as Baidu's DeepSpeech \cite{amodei2016deep} or the Mozilla Speech to text engine\footnote{https://github.com/mozilla/DeepSpeech}.
To better understand the interaction of the CTC loss with CD symbols,
we will describe CTC from the graph transformer point of view.

\begin{figure}
    \centering
    \setlength\tabcolsep{0pt}
    \begin{tabular}{c}
    \includegraphics[width=.45\columnwidth]{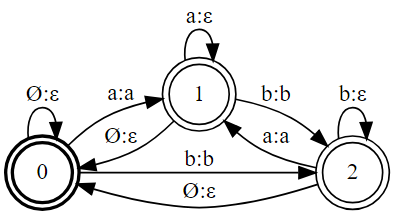} \\
    \textit{(a) Decoding transducer} \\
    \includegraphics[width=1.0\columnwidth]{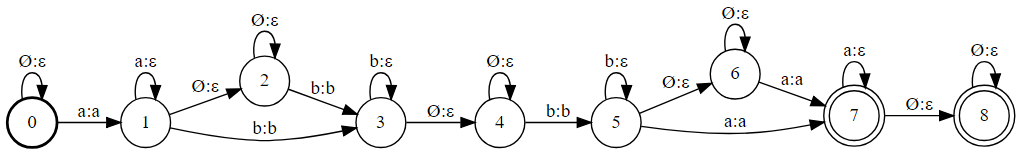} \\
    \textit{(b) Training graph for a single utterance \textit{``abba''}}
    \end{tabular}
    \caption{CTC transducers for a two-letter alphabet $\L = \{a, b\}$}
    \label{fig:ctc_dec_graph}
\end{figure}

\subsection{Locally and globally normalized CTC}
Let $Y$ be a desired transcript, i.e., a sequence of symbols over an alphabet $\L$. An extended transcript $\hat{Y}$ of symbols from the alphabet $\hat{\L}=\L\cup\{\blank\}$  is a longer sequence formed by repeating elements of $Y$ and padding them with a special blank ($\blank$) token. Consecutive elements of $\hat{Y}$ correspond to input frames. Furthermore, $Y$ can be uniquely recovered from $\hat{Y}$ by removing repetitions and blanks, implemented by a function $B(\hat{Y}) = Y$ \cite{graves2006connectionist}. This operation can also be implemented using a Finite State Transducer (FST) \cite{mohri_speech_2008} shown in Figure~\ref{fig:ctc_dec_graph} for a two-letter alphabet $\L = \{a, b\}$.
Equivalently, all extended transcripts that map to $Y_e$ can be captured with a regular expression
{
\setlength{\abovedisplayskip}{3pt}
\setlength{\belowdisplayskip}{3pt}
\[
    \blank^*\spc a\spc a^*\spc \blank^*\spc b\spc b^*\spc \blank\spc 
    \blank^*\spc b\spc b^*\spc \blank^*\spc a\spc a^*\spc \blank^*\spc,
\]
}
where the ${}^*$ operator denotes zero or more repetitions.

The network reads a sequence of $T$ acoustic frames and computes a matrix $O\in\mathbb{R}^{T \times |\hat{\L}|}$ of unnormalized, non-negative scores $O_{t,l}$ for emitting symbol $l$ from frame $t$. 
The unnormalized score of an entire extended transcript $\hat{Y}$ is defined as the product of scores assigned to all its symbols\footnote{Practical applications replace the product of scores with a sum of logarithms for numerical stability.}:
\begin{equation}
    \hat{S}(\hat{Y}) = \prod_t O_{t, \hat{Y}_t}.
\end{equation}

\subsubsection{CTC-G: Globally normalized CTC}
To compute the probability assigned to a transcript $Y$ we normalize the sum of scores of all extended transcripts that map to $Y$ by the sum of scores of all extended transcripts, as advocated in GTN \cite{lecun_gradientbased_1998} and LF-MMI \cite{povey_purely_2016}:
\begin{equation} \label{eq:global_ctc}
    P(Y|X) = \frac{\sum_{\hat{Y} \in B^{-1}(Y)}\hat{S}(\hat{Y})}
                  {\sum_{\hat{Y} \in \hat{\L}^T: \hat{Y} \textrm{is valid}}\hat{S}(\hat{Y})}.
\end{equation}
Both the numerator and the denominator can be computed using the forward algorithm over graphs unrolled over utterance frames: the utterance graph (Figure \ref{fig:ctc_dec_graph}b) for the numerator and the decoding graph (Figure \ref{fig:ctc_dec_graph}a) for the denominator. The time complexity depends on the number of states in the graph. For CD systems it is dominated by the denominator computation.

\subsubsection{CTC: Local normalization in CTC}
CTC avoids denominator computation in \eqref{eq:global_ctc} by using framewise normalization of scores. Observe that for any $Z_t$ we have
\begin{align*}
    P(Y|X) &= \frac{\sum_{\hat{Y} \in B^{-1}(Y)}\hat{S}(\hat{Y})}
                  {\sum_{\hat{Y} \in \hat{\L}^T: \hat{Y} \textrm{is valid}}\hat{S}(\hat{Y})} \\
           &= \frac{\sum_{\hat{Y} \in B^{-1}(Y)}\prod_t O_{t, \hat{Y}_t}/Z_t}
                  {\sum_{\hat{Y} \in \hat{\L}^T: \hat{Y} \textrm{is valid}}\prod_t O_{t, \hat{Y}_t}/Z_t}.
\end{align*}
Validity of an extended transcript refers to the correct overlap of neighboring symbols, which is an issue if CD symbols are used. With ordinary context independent (CI) symbols, every extended transcript is valid.
Taking $Z_t = \sum_y O_{t,y}$ we can locally normalize network outputs into probabilities $p(\hat{Y}_t=y) = O_{t,y}/Z_t$. 
Furthermore, when all extended transcripts are valid, which is the case when context-independent (CI) symbols are used, the denominator $\sum_{\hat{Y} \in \hat{\L}^T}\prod_t p(\hat{Y}_t=y)$ is always $1.0$ and need not be computed recovering the familiar CTC formula
\begin{equation} \label{eq:local_ctc}
    P(Y|X) =  \smashoperator\sum_{\hat{Y} \in B^{-1}(Y)}\prod_t p(\hat{Y}_t|X).
\end{equation}

\subsection{CTC loss with context-dependent symbols}
\label{sec:ctc_cd_syms}
We illustrate the CTC criterion with CD symbols on an example.
Let $[a, b, b, a]$ be the target transcript, which corresponds to the sequence $Y_E = [\blank{}a, ab, bb, ba]$ of bi-charactes with $\blank{}a$, $aa$, and $ba$ denoting CD-variants of $a$. The extended alphabet has 7 symbols (1 blank and 6 CD symbols). All extended transcripts of $Y_E$ match the expression 
{
\setlength{\abovedisplayskip}{3pt}
\setlength{\belowdisplayskip}{3pt}
\[
    \blank^*\spc \blank{}a\spc \blank{}a^*\spc \blank^*\spc ab\spc ab^*\spc
    \blank^*\spc bb\spc bb^*\spc \blank^*\spc ba\spc ba^*\spc \blank^* .
\]
}
Please note that the blank symbol between repeated $b$s became optional, because the two emissions of $b$ have different contexts.

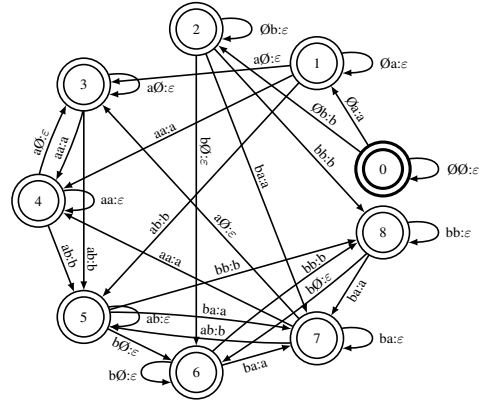
\begin{figure}
    \centering
    \scalebox{0.6}{%

\begin{tikzpicture}[>=latex,line join=bevel,]
  \pgfsetlinewidth{1bp}
\pgfsetcolor{black}
  \draw [->] (221.1bp,154.0bp) .. controls (201.26bp,168.99bp) and (163.75bp,197.33bp)  .. (131.86bp,221.42bp);
  \draw (198.33bp, 178.69bp) node [rotate=-37.07] {Øb:b};
  \draw [->] (64.615bp,55.88bp) .. controls (96.846bp,65.051bp) and (169.32bp,85.672bp)  .. (218.4bp,99.641bp);
  \draw (137.75bp, 82.54bp) node [rotate=15.88]{bb:b};
  \draw [->] (133.96bp,237.91bp) .. controls (143.64bp,239.42bp) and (152.85bp,237.33bp)  .. (152.85bp,231.64bp) .. controls (152.85bp,228.08bp) and (149.25bp,225.92bp)  .. (133.96bp,225.36bp);
  \draw (166.85bp,231.64bp) node {Øb:$\varepsilon$};
  \draw [->] (134.26bp,21.035bp) .. controls (144.01bp,23.813bp) and (156.71bp,27.433bp)  .. (177.48bp,33.356bp);
  \draw (154.84bp, 21.28bp) node [rotate=15.91]{ba:a};
  \draw [->] (64.048bp,57.55bp) .. controls (73.726bp,59.063bp) and (82.937bp,56.971bp)  .. (82.937bp,51.276bp) .. controls (82.937bp,47.716bp) and (79.339bp,45.565bp)  .. (64.048bp,45.002bp);
  \draw (94.937bp,50.276bp) node {ab:$\varepsilon$};
  \draw [->] (250.02bp,110.47bp) .. controls (259.7bp,111.98bp) and (268.91bp,109.89bp)  .. (268.91bp,104.2bp) .. controls (268.91bp,100.64bp) and (265.31bp,98.485bp)  .. (250.02bp,97.922bp);
  \draw (281.41bp,104.2bp) node {bb:$\varepsilon$};
  \draw [->] (124.36bp,216.11bp) .. controls (137.31bp,182.66bp) and (167.94bp,103.57bp)  .. (187.33bp,53.486bp);
  \draw (159.77bp, 139.34bp) node [rotate=-68.83]{ba:a};
  \draw [->] (47.915bp,180.13bp) .. controls (45.293bp,169.91bp) and (40.511bp,156.77bp)  .. (31.146bp,136.78bp);
  \draw (34.68bp, 157.24bp) node [rotate=68.85]{aa:a};
  \draw [->] (35.896bp,130.34bp) .. controls (45.574bp,131.85bp) and (54.785bp,129.76bp)  .. (54.785bp,124.07bp) .. controls (54.785bp,120.51bp) and (51.187bp,118.36bp)  .. (35.896bp,117.79bp);
  \draw (66.285bp,124.07bp) node {aa:$\varepsilon$};
  \draw [->] (176.9bp,34.955bp) .. controls (151.83bp,34.509bp) and (103.91bp,38.634bp)  .. (64.164bp,45.363bp);
  \draw (128.80bp, 42.95bp) node [rotate=-5.27]{ab:b};
  \draw [->] (26.255bp,108.63bp) .. controls (29.913bp,99.173bp) and (34.678bp,86.851bp)  .. (42.475bp,66.691bp);
  \draw (39.95bp, 89.85bp) node [rotate=-68.86]{ab:b};
  \draw [->] (129.48bp,219.42bp) .. controls (149.53bp,197.4bp) and (191.9bp,150.87bp)  .. (223.15bp,116.57bp);
  \draw (198.67bp, 152.21bp) node [rotate=-47.67] {bb:b};
  \draw [->] (64.048bp,203.13bp) .. controls (73.726bp,204.65bp) and (82.937bp,202.56bp)  .. (82.937bp,196.86bp) .. controls (82.937bp,193.3bp) and (79.339bp,191.15bp)  .. (64.048bp,190.59bp);
  \draw (96.437bp,194.86bp) node {aØ:$\varepsilon$};
  \draw [->] (250.02bp,150.21bp) .. controls (259.7bp,151.73bp) and (268.91bp,149.64bp)  .. (268.91bp,143.94bp) .. controls (268.91bp,140.38bp) and (265.31bp,138.23bp)  .. (250.02bp,137.67bp);
  \draw (284.41bp,143.94bp) node {ØØ:$\varepsilon$};
  \draw [->] (178.31bp,45.381bp) .. controls (148.39bp,60.28bp) and (81.208bp,93.733bp)  .. (35.431bp,116.53bp);
  \draw (102.31bp, 87.56bp) node [rotate=-26.47]{aa:a};
  \draw [->] (128.54bp,29.757bp) .. controls (146.75bp,46.964bp) and (184.93bp,76.188bp)  .. (218.81bp,97.975bp);
  \draw (191.63bp, 85.26bp) node [rotate=37.08] {bb:b};
  \draw [->] (208.98bp,216.53bp) .. controls (218.66bp,218.04bp) and (227.87bp,215.95bp)  .. (227.87bp,210.25bp) .. controls (227.87bp,206.69bp) and (224.27bp,204.54bp)  .. (208.98bp,203.98bp);
  \draw (241.37bp,210.25bp) node {Øa:$\varepsilon$};
  \draw [->] (64.909bp,54.205bp) .. controls (89.974bp,54.651bp) and (137.9bp,50.526bp)  .. (177.64bp,43.797bp);
  \draw (129.77bp, 55.95bp) node [rotate=-5.27]{ba:a};
  \draw [->] (182.24bp,198.04bp) .. controls (157.43bp,170.83bp) and (97.391bp,104.97bp)  .. (59.807bp,63.747bp);
  \draw (95.00bp, 110.40bp) node [rotate=47.65]{ab:b};
  \draw [->] (178.31bp,202.75bp) .. controls (148.39bp,187.86bp) and (81.208bp,154.4bp)  .. (35.431bp,131.61bp);
  \draw (98.91bp, 167.57bp) node [rotate=26.47]{aa:a};
  \draw [->] (225.71bp,158.0bp) .. controls (220.38bp,166.62bp) and (213.43bp,177.84bp)  .. (202.06bp,196.21bp);
  \draw (217.71bp, 181.73bp) node [rotate=-58.24]{Øa:a};
  \draw [->] (20.806bp,140.8bp) .. controls (23.428bp,151.01bp) and (28.21bp,164.16bp)  .. (37.576bp,184.15bp);
  \draw (21.32bp, 161.06bp) node [rotate=68.85]{aØ:$\varepsilon$};
  \draw [->] (224.22bp,90.939bp) .. controls (206.01bp,73.732bp) and (167.83bp,44.508bp)  .. (133.95bp,22.721bp);
  \draw (194.60bp, 72.73bp) node [rotate=37.08]{bØ:$\varepsilon$};
  \draw [->] (63.263bp,43.901bp) .. controls (72.347bp,39.382bp) and (84.181bp,33.496bp)  .. (103.54bp,23.864bp);
  \draw (74.66bp, 32.44bp) node [rotate=-26.45]{bØ:$\varepsilon$};
  \draw [->] (208.98bp,44.158bp) .. controls (218.66bp,45.671bp) and (227.87bp,43.579bp)  .. (227.87bp,37.884bp) .. controls (227.87bp,34.324bp) and (224.27bp,32.173bp)  .. (208.98bp,31.61bp);
  \draw (239.87bp,37.884bp) node {ba:$\varepsilon$};
  \draw [->] (103.04bp,22.774bp) .. controls (93.36bp,24.287bp) and (84.15bp,22.195bp)  .. (84.15bp,16.5bp) .. controls (84.15bp,12.94bp) and (87.75bp,10.789bp)  .. (103.04bp,10.226bp);
  \draw (72.85bp,14.5bp) node {bØ:$\varepsilon$};
  \draw [->] (176.75bp,208.72bp) .. controls (151.85bp,206.42bp) and (104.63bp,202.05bp)  .. (64.869bp,198.38bp);
  \draw (162.51bp, 213.16bp) node [rotate=5.28]{aØ:$\varepsilon$};
  \draw [->] (182.24bp,50.095bp) .. controls (157.43bp,77.308bp) and (97.391bp,143.16bp)  .. (59.807bp,184.39bp);
  \draw (136.95bp, 108.55bp) node [rotate=-47.65]{aØ:$\varepsilon$};
  \draw [->] (118.35bp,215.11bp) .. controls (118.35bp,178.12bp) and (118.35bp,88.346bp)  .. (118.35bp,33.08bp);
  \draw (124.08bp, 155.09bp) node [rotate=-90.00]{bØ:$\varepsilon$};
  \draw [->] (225.71bp,90.133bp) .. controls (220.38bp,81.517bp) and (213.43bp,70.292bp)  .. (202.06bp,51.927bp);
  \draw (218.01bp, 68.13bp) node [rotate=58.24]{ba:a};
  \draw [->] (48.437bp,180.16bp) .. controls (48.437bp,155.15bp) and (48.437bp,107.72bp)  .. (48.437bp,67.782bp);
  \draw (53.64bp, 87.62bp) node [rotate=-90.00]{ab:b};
\begin{scope}
  \definecolor{strokecol}{rgb}{0.0,0.0,0.0};
  \pgfsetstrokecolor{strokecol}
  \draw [solid] (193.37bp,210.25bp) ellipse (12.5bp and 12.5bp);
  \draw [solid] (193.37bp,210.25bp) ellipse (16.5bp and 16.5bp);
  \draw (193.37bp,210.25bp) node {1};
\end{scope}
\begin{scope}
  \definecolor{strokecol}{rgb}{0.0,0.0,0.0};
  \pgfsetstrokecolor{strokecol}
  \draw [very thick] (234.41bp,143.94bp) ellipse (12.5bp and 12.5bp);
  \draw [very thick] (234.41bp,143.94bp) ellipse (11.75bp and 11.75bp);
  \draw [very thick] (234.41bp,143.94bp) ellipse (16.5bp and 16.5bp);
  \draw [very thick] (234.41bp,143.94bp) ellipse (17.25bp and 17.25bp);
  \draw (234.41bp,143.94bp) node {0};
\end{scope}
\begin{scope}
  \definecolor{strokecol}{rgb}{0.0,0.0,0.0};
  \pgfsetstrokecolor{strokecol}
  \draw [solid] (48.44bp,196.86bp) ellipse (12.5bp and 12.5bp);
  \draw [solid] (48.44bp,196.86bp) ellipse (16.5bp and 16.5bp);
  \draw (48.437bp,196.86bp) node {3};
\end{scope}
\begin{scope}
  \definecolor{strokecol}{rgb}{0.0,0.0,0.0};
  \pgfsetstrokecolor{strokecol}
  \draw [solid] (118.35bp,231.64bp) ellipse (12.5bp and 12.5bp);
  \draw [solid] (118.35bp,231.64bp) ellipse (16.5bp and 16.5bp);
  \draw (118.35bp,231.64bp) node {2};
\end{scope}
\begin{scope}
  \definecolor{strokecol}{rgb}{0.0,0.0,0.0};
  \pgfsetstrokecolor{strokecol}
  \draw [solid] (48.44bp,51.28bp) ellipse (12.5bp and 12.5bp);
  \draw [solid] (48.44bp,51.28bp) ellipse (16.5bp and 16.5bp);
  \draw (48.437bp,51.276bp) node {5};
\end{scope}
\begin{scope}
  \definecolor{strokecol}{rgb}{0.0,0.0,0.0};
  \pgfsetstrokecolor{strokecol}
  \draw [solid] (20.28bp,124.07bp) ellipse (12.5bp and 12.5bp);
  \draw [solid] (20.28bp,124.07bp) ellipse (16.5bp and 16.5bp);
  \draw (20.285bp,124.07bp) node {4};
\end{scope}
\begin{scope}
  \definecolor{strokecol}{rgb}{0.0,0.0,0.0};
  \pgfsetstrokecolor{strokecol}
  \draw [solid] (193.37bp,37.88bp) ellipse (12.5bp and 12.5bp);
  \draw [solid] (193.37bp,37.88bp) ellipse (16.5bp and 16.5bp);
  \draw (193.37bp,37.884bp) node {7};
\end{scope}
\begin{scope}
  \definecolor{strokecol}{rgb}{0.0,0.0,0.0};
  \pgfsetstrokecolor{strokecol}
  \draw [solid] (118.35bp,16.5bp) ellipse (12.5bp and 12.5bp);
  \draw [solid] (118.35bp,16.5bp) ellipse (16.5bp and 16.5bp);
  \draw (118.35bp,16.5bp) node {6};
\end{scope}
\begin{scope}
  \definecolor{strokecol}{rgb}{0.0,0.0,0.0};
  \pgfsetstrokecolor{strokecol}
  \draw [solid] (234.41bp,104.2bp) ellipse (12.5bp and 12.5bp);
  \draw [solid] (234.41bp,104.2bp) ellipse (16.5bp and 16.5bp);
  \draw (234.41bp,104.2bp) node {8};
\end{scope}
\end{tikzpicture}

     }
    \caption{Bi-character CTC decoding transducer for a two-letter alphabet $\L = \{a, b\}$.
             It imposes overlap between subsequent bi-characters.
             }
    \label{fig:ctc_bi_graph}
\end{figure}

From the example we can see that CD symbols which form an extended transcript overlap with the symbol at frame $t$, setting the context for frame $t+1$. This means that strings over the alphabet $\hat{Y}$ with improper overlaps are invalid and must be removed from the denominator sum in \eqref{eq:global_ctc}. This can be achieved with the forward algorithm applied to a sparsely connected decoding graph, e.g., the bi-character one in Figure~\ref{fig:ctc_bi_graph}. 

However, the locally normalized CTC loss \eqref{eq:local_ctc} implicitly sums over all strings in $\hat{Y}^*$, including the invalid ones whose CD symbols do not overlap. This makes the network prone to overfitting, because it must learn to properly overlap the symbols that are predicted and thus remove all ambiguity from its outputs. In practice this means that CTC predictions will become very sharp, selecting for each frame only one CD-symbol with high probability. The model is forced to start modeling the language in order to be able to essentially output a single hypothesis. During decoding, this internal language model conflicts the external one,
requiring the use of low acoustic model weights (cf. Figure~\ref{fig:ctc_fst_acwt}).

Under a different interpretation, local normalization forces the network to differentiate between the same symbol in different contexts, even though they may correspond to exactly the same sound.
Clustering CD-symbols using decision tress \cite{senior2015context} helps by grouping similarly sounding tied symbols.
However, even with tied symbols, there are many invalid extended transcripts which the network will learn not to emit.

\subsubsection{CTC-GB: context-dependent blanks in CTC-G}
Globally normalized CTC allows to introduce multiple context-dependent blank characters (\textbf{CTC-GB}). This brings the CTC topology closer to a classical tri-state HMM, where blanks serve as secondary subcharacter states. In our running example $Y_E$, all valid extended transcripts with CD blanks fit the pattern
{
\setlength{\abovedisplayskip}{3pt}
\setlength{\belowdisplayskip}{3pt}
\[
    \blank\blank^*\spc \blank{}a\spc \blank{}a^*\spc a\blank^*\spc ab\spc ab^*\spc
    b\blank^*\spc bb\spc bb^*\spc b\blank^*\spc ba\spc ba^*\spc a\blank^* ,
\]
}
where $\blank\blank$, $a\blank$, and $b\blank$ are the CD blank symbols. We found that the CD blanks are especially helpful when we tie the prototypes used by the scoring layer in the network, as described next. 

\section{CD Embeddings for End-to-end Training}
The last parametrized layer of a typical deep neural model
is a linear layer, also called an embedding or a look-up layer.
It
stores a prototype vector per each of $N$ output symbols and has $\mathcal{O}(N)$ parameters.
Because CD targets are strings of $D$ symbols from the alphabet $\hat{\L}$,
the embedding layer naively requires $\mathcal{O}(|\hat{\L}|^D)$ parameters.
This exponential growth with $D$ can be alleviated by fixing the number of tied states,
and mapping each of $|\hat{\L}^D|$ strings to a tied state in the scoring layer with a 
decision tree. %

\begin{figure}[b]
  \centering
  \includegraphics[width=0.7\linewidth]{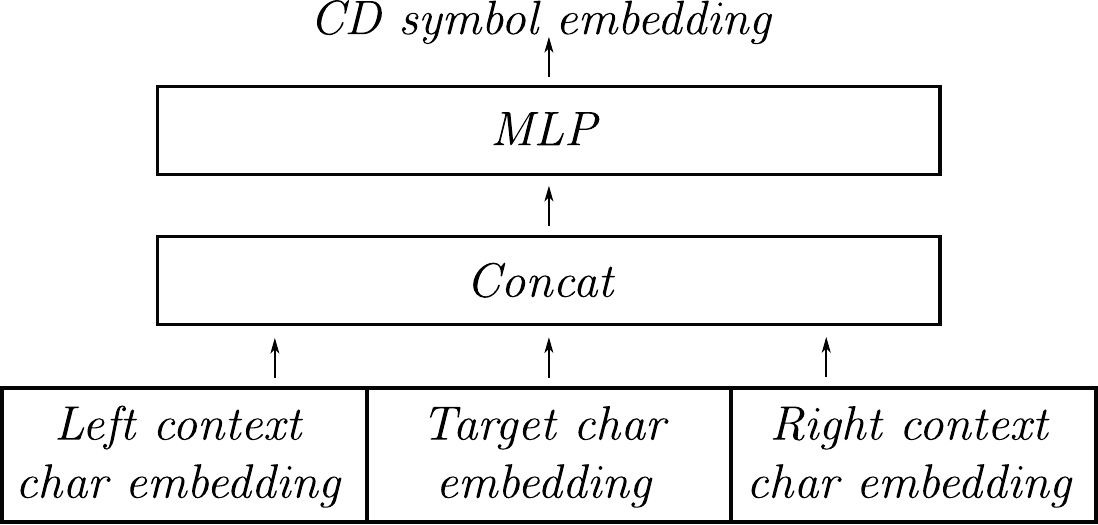}
\caption{Architecture of the Context-Dependent Embedding (CDE) network.
         Embeddings of individual characters are combined to form a CD symbol embedding.}
\label{fig:emb_mlp}
\end{figure}

We propose an alternative way to handle large numbers of output symbols:
keep the full set of output symbols (which still grows exponentially with the size of the context),
but make their prototypes co-dependent by generating them with an auxiliary
Context-Dependent Embedding (CDE) neural network,
analogous to a state-tying decision tree.
CDE reduces the number of parameters in the scoring layer,
enables learning similarities between contexts,
and generalization to previously unseen ones.
Prototypes of $n$-gram CD symbols (in our case bi-chars and tri-chars)
are computed with $n$ separate character embedding layers.
Those embeddings of individual characters are then concatenated
and passed through a ReLU MLP (Figure~\ref{fig:emb_mlp}).

\section{Model Description}
We use the Wall Street Journal dataset,
with the typical split into \textit{Si284}, \textit{Dev93},
and \textit{Eval92} as train, dev and test sets, respectively.
We calculate 80 filterbank features along with the energy in each frame,
extend them with temporal $\Delta$s and $\Delta$-$\Delta$s,
and apply global cepstral mean and variance normalization (CMVN).
We pre-process the data using Kaldi \cite{povey2011kaldi},
implement the neural network in PyTorch \cite{paszke2017automatic} and use Kaldi's FST decoder.

Our model has two convolutional layers with ReLU activations, dimensionality $32$, kernel sizes $7\times 7$, and strides $1\times 2$ and $3 \times 1$, meaning that the first layer halves the resolution along the frequency axis and the second layer reduces the length of the utterance. We apply batch normalization \cite{ioffe_batch_2015} after each convolution. The convolutions are followed by four BiLSTM layers with 320 cells each. Therefore, scoring layer prototype vectors are also 320 dimensional.
The alphabet has $|\hat{\L}|=49$ symbols, which accounts for 2401 bi-characters,
and over $117 k$ possible tri-characters.
We limit the set of tri-characters from $117 k$ to less than $18 k$
which appear in the training data and the language model.
We treat the leftmost and rightmost characters of a tri-char as left and right contexts.
The context-dependent scoring network embeds each CI symbol into 160 dimensions for bi-chars and 110 dimensions for tri-chars, then uses 2 ReLU layers with 320 units and an affine projection into 320 dimensions.

Hyperparameters were adjusted on the baseline CTC context-independent model.
All models are trained with batches of 16 utterances
using Adam optimizer~\cite{kingma2014adam} and
learning rate $0.001$, which is halved every 5 epochs starting from the 32\textsuperscript{nd} epoch. We use Polyak averaging with decay $0.998$ \cite{polyak_acceleration_1992}.
All parameters are initialized using PyTorch defaults.
We regularize with Gaussian weight noise with peak standard deviation $\sigma=0.15$~\cite{graves2011practical},
which increases linearly during initial $20 k$ steps. 

\begin{figure}[t]
    \centering
    \includegraphics[width=1.05\columnwidth]{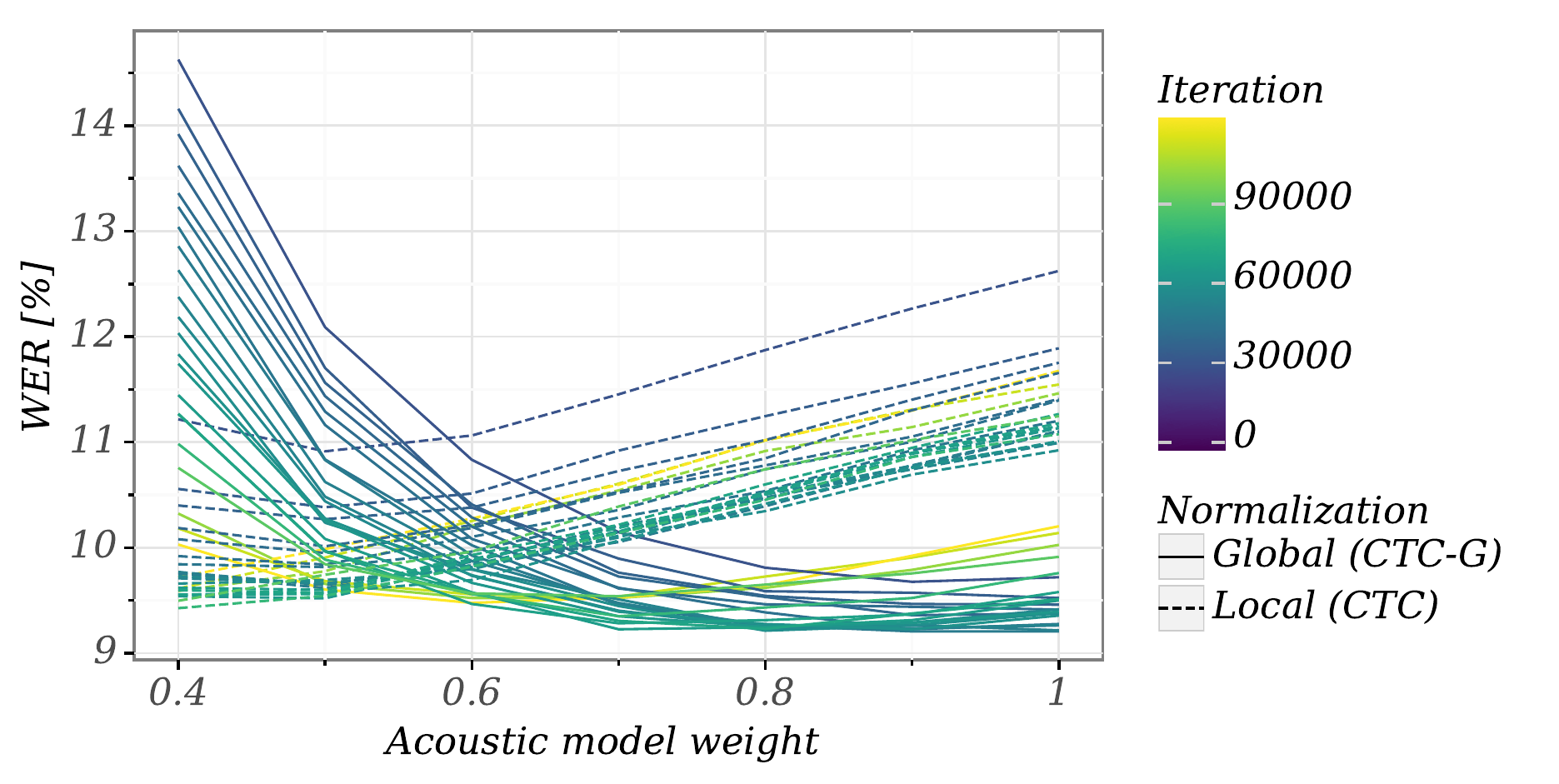}
    \caption{Globally (CTC-G) versus locally normalized CTC with CD symbols under external LM. CTC-G reaches lower WER with higher acoustic weights than CTC, indicating that its outputs are more ambiguous and conflict less with the LM. The range of optimal acoustic weights is also broader for CTC-G.
    }
    \label{fig:ctc_fst_acwt}
\end{figure}

\section{Experiments}
We first confirm that locally normalized CTC with CD symbols conflicts with external language models. Figure~\ref{fig:ctc_fst_acwt} relates \textit{Dev93} decoding accuracy with acoustic weight for global and local normalizations.
For a given symbol, locally normalized CTC sharply differentiates between its contexts.
This has to be mitigated during decoding with low acoustic model weights,
which is equivalent to increasing the temperature of its SoftMax output classifier.
Globally normalized CTC-G reaches a lower overall WER with higher acoustic weights,
indicating that it produces more ambiguous hypotheses, which are resolved by the LM.
We observed a similar trend for other locally and globally normalized models.
In fact, for all CD models in Table~\ref{tab:wsj} the optimal acoustic weight was close to $0.4$ for locally normalized ones and to $1.0$ for globally normalized ones.

\begin{table}[t]
\caption{WER (\%) of character-based models on WSJ, with external LMs for locally normalized CTC, globally normalized CTC-G (e.q.~\eqref{eq:global_ctc}), CTC-GB with CD blanks. CDE denotes networks with the deep CD-symbol embedder.
With the exception of tri-char models, we average 3 independent training runs.
}
\label{tab:wsj}
\centering
\begin{tabular}{l c c c c}
\toprule
      & \multicolumn{2}{c}{Bg LM} & \multicolumn{2}{c}{Tg LM~\cite{miao2015eesen}} \\
Model                 & Dev93 & Eval92 & Dev93 & Eval92 \\
\midrule
LF-MMI~\cite{hadian18end-to-end}  & & & & $\phantom{{}^*}$5.2${}^*$ \\
\multicolumn{2}{l}{LF-MMI Bi-char~\cite{hadian18end-to-end}}  & & & $\phantom{{}^*}$4.1${}^*$ \\
Eesen~\cite{miao2015eesen}    & &     & & 7.3 \\
Gram-CTC~\cite{liu2017gram}   & & & & 6.7 \\
CTC/ASG \cite{zeghidour_endtoend_2018} & & & 9.5 & 6.6 \\
\midrule
 CTC             & 12.1 & 8.7 & 9.3 & 6.6 \\
\midrule
 CTC Bi-char      & 12.0 & 8.4 & 9.4 & 6.4 \\
 CTC Bi. CDE    & 11.9 & 8.4 & 9.3 & 6.4 \\
\midrule
 CTC-G Bi-char    & 11.8 & 8.4 & 9.0 & \textbf{6.2} \\
 CTC-G Bi. CDE  & 11.6 & 8.7 & 9.0 & 6.5 \\
 CTC-GB Bi-char   & 11.8 & 8.5 & 9.2 & 6.5 \\
 CTC-GB Bi. CDE & 11.5 & 8.5 & \textbf{8.8} & \textbf{6.2} \\
\midrule
 CTC Tri-char     & 11.4 & 8.3 & 9.4 & 6.5 \\
 CTC Tri. CDE     & \textbf{11.3} & \textbf{8.2} & 8.9 & 6.4 \\
\bottomrule
\end{tabular}
{\small ${}^*$ Uses additional data augmentation}
\end{table}

We report in Table~\ref{tab:wsj} error rates for CI and CD variants of CTC
using the standard bigram (\textit{Bg}) and trigram (\textit{Tg})~\cite{miao2015eesen} language models.
Early stopping checkpoints and acoustic weights for \textit{Eval92} have been selected
by best WER on \textit{Dev93}. We average the reported scores over 3 independent runs.
As in previous sections,
\textit{CTC} denotes the typical, locally normalized variant \eqref{eq:local_ctc},
\textit{CTC-G} the globally normalized one \eqref{eq:global_ctc}, and \textit{CTC-GB}
the globally normalized one with context-dependent blanks.
Additionally, CDE denotes cases for which CD symbol prototypes were computed
using the auxiliary embedding network instead of a look-up table of prototypes.
Our baselines match the results of other CTC implementations \cite{miao2015eesen,liu2017gram,zeghidour_endtoend_2018}.
For reference we provide LF-MMI results,
however not directly comparable due to their data augmentation techniques.

The gains reported on the \textit{Dev93} set, however small, are consistent
and match the conclusions of our theoretical CTC loss analysis.
Locally normalized bi-character CTC yields about $0.2$ percentage point WER reduction, with slight gains coming from using CDE.
Globally normalized CD CTC models reach the best performance, improving upon the baseline by $0.4$ percentage points.
However, the CDE module brings mixed results for globally normalized bi-character models. It yields a small reduction of the number of parameters,
improves the error rates when contextual blanks are used,
but seems inferior to a simple prototype look-up table under the typical CTC topology. This may indicate that a global CTC blank should have a representation that is unique and separate from other CD symbols, while the contextual blanks benefit from sharing of their prototypes.

The benefits of using the CDE module with tri-characters are more apparent.
Not only it yields lower error rates, reaching the best WER with the bigram language model,
but also the CDE can handle tri-characters not seen during training,
and requires substantially fewer parameters. In our case CDE has about $324k$ parameters ($320k$ versus $5.5M$ required for prototypes of all $17k$ tri-characters allowed by the \emph{Tg} language model).

We attribute the poor performance of the tri-character CTC %
when decoded with the trigram LM,
to overfitting the internal language model as described in Section~\ref{sec:ctc_cd_syms}. In fact, many of the triphones allowed by the trigram LM are not present in the training set, and the locally normalized CTC loss will make them improbable despite their prototypes being tied to other CD symbols by the CDE module, which yields only a small improvement over the CI baseline. Unfortunately, our forward-backward implementation does not allow computing the forward cost of the denominator graph in \eqref{eq:global_ctc} with 17k states and we were not able to build globally normalized tri-character CTC models.

\section{Related Work}
CD symbols in neural ASR systems improve performance of CTC \cite{senior2015context} and  1-state ``chain'' HMM models \cite{hadian18end-to-end}, with best results obtained with decision trees that map contexts to network targets. In \cite{senior2015context} this tree is built from activations of the neural network, while \cite{hadian18end-to-end} used a full bigram output which was slightly inferior to a decision tree obtained using a HMM model. The CDE module aims to recover the benefits of context-dependent outputs, but in a fully neural model that is trainable from scratch.

Several authors have proposed to use multicharacter, or multiphoneme output tokens to reduce the number of emissions. The DeepSpeech 2 model used non-overlappling character bigrams \cite{amodei2016deep}, while \cite{liu2017gram} and \cite{siohan2017ctc} dynamically chose a decomposition of the output sequence. This idea was also explored in \cite{chan2016latent} in the context of sequence-to-sequence models. We do not aim to reduce the length of target sequences, but to enable expressing of dependency of symbol emissions on their context.

Perhaps most similar to our work is \cite{yadav2017deep} in which a neural network is trained to replace a state-tying decision tree. However, this implies a multistage training procedure in which both classical GMM-HMM and neural systems are built. In contrast, we strive to keep a simple, one stage training procedure.

Finally, Hypernetworks \cite{ha2017hypernetworks} expand on the idea of generating weights of neural network using an auxiliary module. We employ this idea in the CDE module and generate the prototypes for context-dependent symbols.

\section{Conclusions}

We have analyzed from a theoretical and experimental viewpoint the behavior of CTC with context-dependent targets. We have identified and addressed two major shortcomings: overfitting of locally normalized CTC, and the expansion of the number of parameters in the final layer of the network. Our proposed solutions are global normalization of the loss and dynamical computation of the final weight matrix using an auxiliary neural network, which allowed us to train compact networks with tri-character outputs. The changes preserve the simplified training procedure valued in the neural end-to-end systems, while yielding a 6\% relative WER improvement on the WSJ dataset.

The scope for future work includes better integration with language models,
scaling the global normalization to larger contexts, and experiments with LF-MMI chain models.
We would also like to test the method on larger datasets.

\bibliographystyle{IEEEtran}
\bibliography{refs}

\begin{thebibliography}{10}
\providecommand{\url}[1]{#1}
\csname url@samestyle\endcsname
\providecommand{\newblock}{\relax}
\providecommand{\bibinfo}[2]{#2}
\providecommand{\BIBentrySTDinterwordspacing}{\spaceskip=0pt\relax}
\providecommand{\BIBentryALTinterwordstretchfactor}{4}
\providecommand{\BIBentryALTinterwordspacing}{\spaceskip=\fontdimen2\font plus
\BIBentryALTinterwordstretchfactor\fontdimen3\font minus
  \fontdimen4\font\relax}
\providecommand{\BIBforeignlanguage}[2]{{%
\expandafter\ifx\csname l@#1\endcsname\relax
\typeout{** WARNING: IEEEtran.bst: No hyphenation pattern has been}%
\typeout{** loaded for the language `#1'. Using the pattern for}%
\typeout{** the default language instead.}%
\else
\language=\csname l@#1\endcsname
\fi
#2}}
\providecommand{\BIBdecl}{\relax}
\BIBdecl

\bibitem{hoshen2015speech}
Y.~Hoshen, R.~J. Weiss, and K.~W. Wilson, ``Speech acoustic modeling from raw
  multichannel waveforms,'' in \emph{2015 IEEE International Conference on
  Acoustics, Speech and Signal Processing (ICASSP)}, 2015, pp. 4624--4628.

\bibitem{amodei2016deep}
D.~Amodei, S.~Ananthanarayanan, and R.~e.~a. Anubhai, ``Deep speech 2:
  End-to-end speech recognition in english and mandarin,'' in \emph{Proceedings
  of the 33rd International Conference on Machine Learning}, 2016, pp.
  173--182.

\bibitem{senior2015context}
A.~Senior, H.~Sak, and I.~Shafran, ``Context dependent phone models for lstm
  rnn acoustic modelling,'' in \emph{2015 IEEE International Conference on
  Acoustics, Speech and Signal Processing (ICASSP)}, 2015, pp. 4585--4589.

\bibitem{hadian18end-to-end}
H.~Hadian, H.~Sameti, D.~Povey, and S.~Khudanpur, ``End-to-end speech
  recognition using {L}attice-free {MMI},'' in \emph{Proceedings of
  Interspeech}, 2018, pp. 12--16.

\bibitem{graves2006connectionist}
A.~Graves, S.~Fern{\'a}ndez, F.~Gomez, and J.~Schmidhuber, ``Connectionist
  temporal classification: labelling unsegmented sequence data with recurrent
  neural networks,'' in \emph{Proceedings of the 23rd International Conference
  on Machine learning}, 2006, pp. 369--376.

\bibitem{rabiner1989tutorial}
L.~R. Rabiner, ``A tutorial on hidden markov models and selected applications
  in speech recognition,'' \emph{Proceedings of the IEEE}, vol.~77, no.~2, pp.
  257--286, 1989.

\bibitem{gales2007application}
M.~Gales and S.~Young, ``The application of hidden markov models in speech
  recognition,'' \emph{Found. Trends Signal Process.}, vol.~1, no.~3, pp.
  195--304, 2007.

\bibitem{young1994tree}
S.~J. Young, J.~J. Odell, and P.~C. Woodland, ``Tree-based state tying for high
  accuracy acoustic modelling,'' in \emph{Proceedings of the Workshop on Human
  Language Technology}, 1994, pp. 307--312.

\bibitem{miao2015eesen}
Y.~Miao, M.~Gowayyed, and F.~Metze, ``Eesen: End-to-end speech recognition
  using deep rnn models and wfst-based decoding.'' in \emph{Proceedings of
  ASRU}, 2015, pp. 167--174.

\bibitem{povey_purely_2016}
D.~Povey, V.~Peddinti, D.~Galvez, P.~Ghahremani, V.~Manohar, X.~Na, Y.~Wang,
  and S.~Khudanpur, ``Purely {Sequence}-{Trained} {Neural} {Networks} for {ASR}
  {Based} on {Lattice}-{Free} {MMI},'' Sep. 2016, pp. 2751--2755.

\bibitem{lecun_gradientbased_1998}
Y.~LeCun, L.~Bottou, Y.~Bengio, and P.~Haffner, ``Gradient-based learning
  applied to document recognition,'' \emph{Proceedings of the IEEE}, vol.~86,
  no.~11, pp. 2278--2324, 1998.

\bibitem{collobert_wav2letter_2016}
R.~Collobert, C.~Puhrsch, and G.~Synnaeve, ``Wav2letter: an {End}-to-{End}
  {ConvNet}-based {Speech} {Recognition} {System},'' \emph{arXiv:1609.03193
  [cs]}, Sep. 2016, arXiv: 1609.03193.

\bibitem{chorowski_attentionbased_2015a}
J.~K. Chorowski, D.~Bahdanau, D.~Serdyuk, K.~Cho, and Y.~Bengio,
  ``Attention-{Based} {Models} for {Speech} {Recognition},'' in \emph{Advances
  in {Neural} {Information} {Processing} {Systems} 28}, C.~Cortes, N.~D.
  Lawrence, D.~D. Lee, M.~Sugiyama, and R.~Garnett, Eds.\hskip 1em plus 0.5em
  minus 0.4em\relax Curran Associates, Inc., 2015, pp. 577--585.

\bibitem{chan_listen_2015}
W.~Chan, N.~Jaitly, Q.~V. Le, and O.~Vinyals, ``Listen, {Attend} and {Spell},''
  \emph{arXiv:1508.01211 [cs, stat]}, Aug. 2015, arXiv: 1508.01211.

\bibitem{mohri_speech_2008}
M.~Mohri, F.~Pereira, and M.~Riley, ``Speech {Recognition} with {Weighted}
  {Finite}-{State} {Transducers},'' in \emph{Springer {Handbook} of {Speech}
  {Processing}}, P.~J.~B. Dr, P.~M.~M. Sondhi, and P.~Y. A.~H. Dr, Eds.\hskip
  1em plus 0.5em minus 0.4em\relax Springer Berlin Heidelberg, Jan. 2008, pp.
  559--584.

\bibitem{povey2011kaldi}
D.~Povey, A.~Ghoshal, G.~Boulianne, L.~Burget, O.~Glembek, N.~Goel,
  M.~Hannemann, P.~Motlicek, Y.~Qian, P.~Schwarz \emph{et~al.}, ``The {K}aldi
  speech recognition toolkit,'' in \emph{IEEE 2011 workshop on automatic speech
  recognition and understanding}, no. EPFL-CONF-192584.\hskip 1em plus 0.5em
  minus 0.4em\relax IEEE Signal Processing Society, 2011.

\bibitem{paszke2017automatic}
A.~Paszke, S.~Gross, S.~Chintala, G.~Chanan, E.~Yang, Z.~DeVito, Z.~Lin,
  A.~Desmaison, L.~Antiga, and A.~Lerer, ``Automatic differentiation in
  {P}y{T}orch,'' 2017.

\bibitem{ioffe_batch_2015}
S.~Ioffe and C.~Szegedy, ``Batch {Normalization}: {Accelerating} {Deep}
  {Network} {Training} by {Reducing} {Internal} {Covariate} {Shift},''
  \emph{arXiv:1502.03167 [cs]}, Feb. 2015, arXiv: 1502.03167.

\bibitem{kingma2014adam}
D.~P. Kingma and J.~Ba, ``Adam: A method for stochastic optimization,''
  \emph{arXiv preprint arXiv:1412.6980}, 2014.

\bibitem{polyak_acceleration_1992}
B.~T. Polyak and A.~B. Juditsky, ``Acceleration of stochastic approximation by
  averaging,'' \emph{SIAM Journal on Control and Optimization}, vol.~30, no.~4,
  pp. 838--855, 1992.

\bibitem{graves2011practical}
A.~Graves, ``Practical variational inference for neural networks,'' in
  \emph{Advances in Neural Information Processing Systems 24}, 2011, pp.
  2348--2356.

\bibitem{liu2017gram}
H.~Liu, Z.~Zhu, X.~Li, and S.~Satheesh, ``{G}ram-{CTC}: Automatic unit
  selection and target decomposition for sequence labelling,'' in
  \emph{Proceedings of the 34th International Conference on Machine Learning},
  2017, pp. 2188--2197.

\bibitem{zeghidour_endtoend_2018}
N.~Zeghidour, N.~Usunier, G.~Synnaeve, R.~Collobert, and E.~Dupoux,
  ``End-to-{End} {Speech} {Recognition} {From} the {Raw} {Waveform},''
  \emph{arXiv:1806.07098 [cs, eess]}, Jun. 2018, arXiv: 1806.07098.

\bibitem{siohan2017ctc}
O.~Siohan, ``{CTC} training of multi-phone acoustic models for speech
  recognition,'' in \emph{Proceedings of Interspeech}, 2017, pp. 709--713.

\bibitem{chan2016latent}
W.~Chan, Y.~Zhang, Q.~Le, and N.~Jaitly, ``Latent sequence decompositions,'' in
  \emph{Proceedings of the International Conference on Learning Representations
  (ICLR)}, 2017.

\bibitem{yadav2017deep}
M.~Yadav and V.~Tyagi, ``Deep triphone embedding improves phoneme
  recognition,'' \emph{arXiv preprint arXiv:1710.07868}, 2017.

\bibitem{ha2017hypernetworks}
D.~Ha, A.~Dai, and Q.~Le, ``Hypernetworks,'' in \emph{Proceedings of the
  International Conference on Learning Representations (ICLR)}, 2017.

\end{thebibliography}

\end{document}